\documentclass[conference]{IEEEtran}

\usepackage{booktabs}
\usepackage{amsmath}
\usepackage{graphicx}
\usepackage{algorithm}
\usepackage{algorithmic}
\usepackage{hyperref}
\usepackage[numbers]{natbib}
\usepackage{dblfloatfix}
\usepackage{balance}
\usepackage{flushend}
\usepackage{pgfplots}
\pgfplotsset{compat=1.18}

\hypersetup{
    colorlinks=true,
    linkcolor=black,
    citecolor=black,
    urlcolor=black,
    pdfauthor={Sriram Loganathan et al.},
    pdftitle={Sparse PPMI Graph Averaging for Random Indexing Embeddings}
}

\title{Sparse PPMI Graph Averaging for Random Indexing Embeddings}

\author{
\IEEEauthorblockN{Sriram Loganathan, Gokul Anand, Aung Bo Bo, Yourui Shao, William B. Andreopoulos}
\IEEEauthorblockA{Department of Computer Science\\
San Jose State University\\
San Jose, CA, USA}
}

\begin{document}

\maketitle

\begin{abstract}
We study a specific sparse post-processing pipeline for Random Indexing (RI) on kinship analogies in a small fairytales corpus. The published artifacts use uniform RI context accumulation with 200 dimensions and eight nonzeros, followed by one residual graph average, $\mathbf{E}=(1-\alpha)\mathbf{E}_0+\alpha\mathbf{P}\mathbf{E}_0$, where $\mathbf{P}$ is a row-normalized PPMI graph and $\alpha=0.3$. Terminal row normalization and per-dimension median/IQR scaling are then applied. On the Google analogy benchmark's family section, 272 of 506 questions are valid for every seed. Across five paired seeds, the complete pipeline raises accuracy from 19.41\% to 30.74\%, a gain of 11.32 percentage points with a nested-bootstrap 95\% confidence interval of [6.93, 15.89]. Robust scaling alone contributes 3.24 points [1.25, 5.38], while graph averaging without robust scaling contributes 6.18 points [2.63, 9.92]. A separate 40-question general grid does not support a general improvement: the full pipeline changes accuracy by -6.00 points [-13.50, -0.50], and averaging without robust scaling changes it by -6.50 points [-14.50, -0.50]. The supported positive claim is therefore limited to the covered fairytales kinship analogy set; the results do not establish a generally effective embedding method.
\end{abstract}


\section{Introduction}

Word embeddings are a foundational component of natural language processing, representing words as dense vectors that capture semantic relationships. Standard methods for constructing these representations fall into two families: count-based methods that factorize large co-occurrence matrices, and predictive methods that train neural models through gradient descent. Both families have produced strong results, but both also require either global matrix operations or repeated parameter updates over large vocabularies, making them expensive in memory and compute.

Sparse graph-based methods provide a narrower alternative. They avoid dense co-occurrence matrix materialization, dense factorization, and gradient training, but they still construct sparse global corpus statistics such as document frequency, co-occurrence marginals, and Positive Pointwise Mutual Information (PPMI) weights. The practical appeal is sparse computation and reproducibility, not strict locality or maximal accuracy.

This paper investigates one such pipeline. Word vectors are initialized from co-occurrence encodings and refined by one weighted neighborhood average on a sparse graph. The pipeline requires no dense matrix decomposition, no backpropagation, and no gradient-trained parameters, but it still has hyperparameters including context window size, embedding dimension, random-index sparsity, PMI shift $k$, top-$K$ pruning, minimum count, smoothing weight $\alpha$, and normalization. The central question is narrow: does this one-step PPMI average improve a uniform RI initialization on the covered fairytales kinship analogies, and which part of the pipeline accounts for the change?

The results support a positive claim only on the covered family category. Across five seeds, the one-step PPMI pipeline improves RI by 11.32 percentage points [6.93, 15.89], and averaging remains positive when robust scaling is removed. A separate general-question grid is negative, so the evidence does not support a broad claim about RI, graph post-processing, or semantic embeddings. Bloom filter initialization is included as a negative result for the tested configuration because its OR-style accumulation may discard frequency information that Random Indexing preserves.

Experiments use the Google analogy test set~\cite{mikolov2013analogies}, SimLex-999~\cite{hill2015}, and WordSim-353~\cite{finkelstein2001} on a fairytales corpus and a Wikipedia-derived text8 sub-corpus. Section II reviews related work. Section III describes the datasets. Section IV presents the method. Section V describes the evaluation protocol. Section VI reports results. Section VII discusses limitations. Section VIII concludes.

\section{Related Work}

\subsection{Word Embeddings}

Word2Vec~\cite{mikolov2013} introduced two influential architectures for predictive word embedding. CBOW predicts a target word from its surrounding context by averaging context vectors, while Skip-gram predicts context words from a target word. Both are trained with negative sampling and have become standard baselines. GloVe~\cite{pennington2014} is a count-based method that combines global co-occurrence statistics with a weighted least-squares objective.

The relationship between count-based and predictive methods has been studied extensively. Baroni et al.~\cite{baroni2014} found that predictive models outperformed count-based methods on many intrinsic tasks. Levy and Goldberg~\cite{levy2014} showed that Skip-gram with negative sampling implicitly factorizes a shifted pointwise mutual information matrix, connecting the two families mathematically. Levy et al.~\cite{levy2015} further showed that many performance differences between methods are explained by hyperparameter choices rather than fundamental architectural differences.

\subsection{Sparse and Memory-Efficient Methods}

Random Indexing~\cite{kanerva2000,sahlgren2005} is the most relevant sparse baseline for this work. It assigns each word a sparse random vector and accumulates context evidence additively by adding the random vectors of neighboring words. This additive property preserves frequency information: a context word that appears 100 times near a target has 100 times more influence than one that appears once. This is a key contrast with Bloom filter OR-based accumulation, which is monotone and does not preserve frequency.

Bloom-based and hash-based embedding ideas have appeared in prior work primarily for compression purposes. Serra and Karatzoglou~\cite{serra2018} used Bloom-style hashing to build compact embedding tables for sparse binary inputs in recommender systems. Svenstrup et al.~\cite{svenstrup2017} proposed hash embeddings that share parameters across vocabulary entries to reduce memory. These methods optimize for memory reduction rather than semantic geometry, which is a different objective from ours.

Recent work on matmul-free language modeling~\cite{zhu2024} has explored eliminating matrix multiplications from transformer architectures entirely. Randomized linear algebra methods~\cite{halko2011} have demonstrated that many machine learning computations can be performed using compressed approximate representations. These directions further motivate asking how far sparse, non-gradient embedding construction can go without dense factorization.

\subsection{Graph Averaging}

Graph-based post-processing methods are closely related to the averaging stage. Retrofitting~\cite{faruqui2015} adjusts pre-trained word vectors to conform to an external lexical knowledge graph. The filter studied here instead uses a corpus-derived sparse graph and applies one residual average. The scalar $\alpha$ is the smoothing weight and $1-\alpha$ is the weight retained on the original RI features.

Post-processing techniques that improve the geometry of learned embeddings are also relevant. Mu and Viswanath~\cite{mu2018} demonstrated that discarding the top principal components of embedding matrices improves performance on similarity and analogy tasks by mitigating anisotropy. The per-dimension robust scaling employed here addresses a related phenomenon: after graph averaging, individual coordinates can exhibit heavy-tailed distributions that distort cosine-based similarity.

\section{Datasets}

\subsection{Fairytales Corpus}

The fairytales corpus is derived from the Fairy Tales from Around the World collection on Kaggle (sourced from Project Gutenberg).\footnote{Fairy Tales from Around the World, Kaggle dataset, \url{https://www.kaggle.com/datasets/annbengardt/fairy-tales-from-around-the-world} (accessed May 1, 2024).} After tokenization and lemmatization with spaCy it contains 1,383,029 tokens and a vocabulary of 18,254 distinct words spanning 149,860 sentence units.

This corpus is small and domain-specific. Semantic analogy coverage is only 272 of 8,869 questions (3.1\%), all from the Google benchmark's family category. Semantic accuracy estimates therefore have high variance and narrow scope. A change of one correctly answered question corresponds to approximately 0.4 percentage points.

\subsection{Text8 Sub2m Corpus}

The text8 sub2m corpus is obtained by subsampling the standard text8 Wikipedia excerpt to approximately two million tokens.\footnote{The text8 corpus is the first 100\,MB of a cleaned English Wikipedia dump, lowercased and reduced to 27 character types, distributed by M. Mahoney at \url{http://mattmahoney.net/dc/textdata.html} (accessed May 1, 2024). The Sub2m subset used here consists of the first approximately two million tokens of text8.} After identical preprocessing it contains 2,000,007 tokens and a vocabulary of 20,181 words. Analogy coverage is substantially higher: 8,946 of 10,675 syntactic questions (83.8\%) and 506 of 8,869 semantic questions (5.7\%).

The cross-corpus comparison is confounded by sentence units. Fairytales averages 9.2 tokens per sentence unit, while text8 averages 46.9 tokens per unit and has no true sentence boundaries. Since context windows are clipped at these units, co-occurrence density differs systematically across corpora.

\subsection{Corpus Statistics}

Semantic coverage and syntactic coverage report the fraction of the Google analogy test set that is answerable given each corpus vocabulary: an analogy question is covered only if all four of its words appear in the vocabulary. These counts are computed by us during evaluation rather than supplied with the corpora, and the evaluation procedure is described in Section~\ref{sec:evaluation}. Coverage is low for the fairytales corpus because its small vocabulary excludes many of the geographic and grammatical terms used in the Google questions.

\begin{table}[H]
\centering
\caption{Corpus statistics and analogy benchmark coverage.}
\label{tab:corpus_stats}
\begin{tabular}{lcc}
\toprule
 & Fairytales & Text8 sub2m \\
\midrule
Tokens & 1,383,029 & 2,000,007 \\
Vocabulary size & 18,254 & 20,181 \\
Sentence units & 149,860 & 42,608 \\
Semantic coverage & 272/8,869 (3.1\%) & 506/8,869 (5.7\%) \\
Syntactic coverage & 2,006/10,675 (18.79\%) & 8,946/10,675 (83.8\%) \\
\bottomrule
\end{tabular}
\end{table}

The fairytales syntactic coverage value in Table~\ref{tab:corpus_stats} is independent of the 18.8\% seed-42 Random Indexing semantic accuracy reported later; the former is a coverage ratio and the latter is 51 correct answers among 272 valid semantic questions.

\section{Method}

\subsection{Overview}

The pipeline comprises three stages: initialization, one-step graph averaging, and terminal normalization. The decomposition in Section VI isolates the contributions of averaging and robust scaling. No dense co-occurrence matrix is materialized, and no dense factorization or gradient training is performed. The method still constructs sparse global co-occurrence statistics and a sparse graph operator. Working memory scales linearly with vocabulary size and embedding dimension plus graph edges.

\subsection{Neighborhood Construction}

For graph construction, each occurrence of a target word $w$ collects context words inside a symmetric window of radius $r=8$ clipped at sentence boundaries. The resulting co-occurrence information is stored as a sparse directed matrix. The graph is directed because the weighting scheme conditions on the target word. Table~\ref{tab:baselines} lists the window used by each reported artifact; the reported RI+PPMI result combines a Random Indexing initialization built with window 10 and a PPMI graph built with window 8.

\subsection{Target-Conditioned Edge Weighting}

Co-occurrence edges are weighted using a target-conditioned scheme that combines local frequency with a global rarity term. For each target word $w$, a pseudo-document is formed by aggregating all context windows centered on $w$. For a neighbor $n$ of $w$, the weight is defined as

\[
T[w,n] = \text{TF}[w,n] \times \text{IDF}[n],
\]

where

\[
\text{TF}[w,n] = \frac{\text{count}(n \text{ in contexts of } w)}{|\text{contexts of } w|}
\]

and

\[
\text{IDF}[n] = \log\left(\frac{1 + |\mathcal{V}|}{1 + \text{DF}[n]}\right) + 1.
\]

Here $\text{DF}[n]$ is the number of distinct target words that have $n$ as a neighbor at least once, and $|\mathcal{V}|$ is the vocabulary size. This follows the standard smoothed-IDF form used in scikit-learn-style TF-IDF implementations~\cite{pedregosa2011}. The smoothed IDF term guarantees strictly positive weights. The resulting graph is directed because the weight $T[w,n]$ is conditioned on the target $w$.

For the PPMI graph, the implementation counts directed co-occurrences $c(w,n)$ inside the same clipped window, discards pairs with count below two, and estimates
\[
\text{PMI}_k[w,n] =
\log\left(\frac{c(w,n) C}{c(w,*)c(*,n)}\right) - \log k,
\]
where $C=\sum_{w,n} c(w,n)$, $c(w,*)$ is the target row count, and $c(*,n)$ is the context column count. The reported PPMI runs use $k=1$, so there is no shift beyond positive truncation. Edge weights are $\text{PPMI}[w,n]=\max(\text{PMI}_k[w,n],0)$, followed by top-$K$ row pruning with $K=50$. In that configuration, $T[w,n]$ in the update equations denotes the pruned PPMI edge weight rather than TF-IDF.

\subsection{Bloom Filter Initialization}

Each target word is assigned an all-zero bit vector of length $m$, following the Bloom filter data structure~\cite{bloom1970}. For each of its top-20 context neighbors (ranked by edge weight), $h$ hash functions are applied using MurmurHash3 with seeds $0, \dots, h-1$, and the corresponding bit positions modulo $m$ are set. Bits are recoded from $\{0,1\}$ to $\{-1,+1\}$ prior to graph averaging. The experiments use $m=200$ bits and $h=5$ hash functions.

In the corrected tested configuration, Bloom initialization still underperforms Random Indexing (Table~\ref{tab:bloom_vs_ri}).\footnote{A previous Bloom implementation applied MurmurHash3 to target strings instead of neighbors and handled negative hashes incorrectly, producing all-zero vectors for much of the vocabulary. The corrected results are reported here as a reproducibility note rather than as part of the method definition.}

\subsection{Random Indexing Initialization}

Each word is assigned a sparse random vector of dimension $d=200$ containing exactly eight nonzero entries drawn uniformly from $\{-1, +1\}$. For every occurrence of a target word, the random vectors of its context neighbors are added without TF-IDF weights. This uniform accumulation is the procedure that produced all five reported RI seed artifacts. Earlier labels calling these artifacts ``RI TF-IDF'' were incorrect. The additive scheme preserves frequency information, in contrast to the monotone OR accumulation used by Bloom filters. Because RI already injects co-occurrence evidence through neighbor-vector accumulation, the later PPMI graph step injects co-occurrence evidence a second time; the result is specific to this initialization and evaluation setting.

\subsection{One-Step Embedding Update}

Let $\mathbf{E}_0$ denote the RI initialization. For each word $w$, the graph average is

\[
\mathbf{M}[w] = \frac{\sum_{n \in N(w)} T[w,n] \cdot \mathbf{E}_0[n]}{\sum_{n \in N(w)} T[w,n]},
\]

where $N(w)$ is the neighbor set and $T[w,n]$ is the pruned PPMI weight. Thus $\mathbf{M}=\mathbf{P}\mathbf{E}_0$, where each nonempty row of the sparse operator $\mathbf{P}$ is divided by the sum of its PPMI weights. The published operator applies the residual average exactly once:

\[
\mathbf{E} = (1-\alpha)\mathbf{E}_0 + \alpha\mathbf{P}\mathbf{E}_0,
\qquad \alpha=0.3.
\]

Words with no retained neighbors have zero rows in $\mathbf{P}$ and retain the $(1-\alpha)\mathbf{E}_0$ component. This equation, followed by the terminal normalization below, is the operator that produced the reported 30.7\% checkpoint.

\begin{algorithm}[t!]
\footnotesize
\caption{One-Step PPMI Graph Average Used for the Headline Runs}
\label{alg:graph_average}
\begin{algorithmic}
\REQUIRE Uniform RI vectors $\mathbf{E}_0$, PPMI operator $\mathbf{P}$, smoothing weight $\alpha=0.3$
\ENSURE Final embeddings $\widehat{\mathbf{E}}$
\STATE $\mathbf{M} \gets \mathbf{P}\mathbf{E}_0$
\STATE $\mathbf{E} \gets (1-\alpha)\mathbf{E}_0 + \alpha\mathbf{M}$
\STATE $\widehat{\mathbf{E}} \gets \text{RobustNormalize}(\mathbf{E})$
\RETURN $\widehat{\mathbf{E}}$
\end{algorithmic}
\end{algorithm}

Let $|\mathcal{V}|$ be the vocabulary size, $E$ the number of directed graph edges, and $d$ the embedding dimension. The sparse multiplication touches each retained edge and accumulates a $d$-dimensional vector, giving $O(E d)$ time. Row normalization costs $O(|\mathcal{V}|d)$, while exact per-coordinate median and percentile computation adds a quantile cost across $|\mathcal{V}|$ values for each dimension. Working memory is $O(|\mathcal{V}|d + E)$ when the graph is stored sparsely. With top-$K$ pruning, $E \leq K|\mathcal{V}|$.

\subsection{Normalization}

The normalization operator used before evaluation is applied in the following order:
\begin{enumerate}
\item Row-wise $\ell_2$ normalization:
\[
\mathbf{v}_w \gets \frac{\mathbf{v}_w}{\|\mathbf{v}_w\|_2 + \epsilon}.
\]

\item Per-dimension robust scaling. For each coordinate $j$, compute the median and interquartile range (IQR) across all word vectors and apply
\[
v_{w,j} \gets \frac{v_{w,j} - \text{median}_j}{\text{IQR}_j + \epsilon}.
\]

\end{enumerate}

The two-stage procedure is denoted by $\text{RobustNormalize}(\cdot)$ in Algorithm~\ref{alg:graph_average}. There is no second row normalization after robust scaling. The decomposition in Section VI reports robust scaling alone and one-step averaging with robust scaling removed.

\subsection{Smoothing Weight}

The single parameter $\alpha \in [0,1]$ controls the balance between the original RI vectors and their graph average. The value $\alpha=0$ exactly reproduces the initialization-only branch, while $\alpha=1$ keeps only the graph average before terminal normalization. The headline configuration uses $\alpha=0.3$. A clean held-out sweep over $\{0,0.15,0.3,0.5,0.7,1\}$ selected 0.3 on 17 development questions that do not overlap the reported family or general question sets.

\subsection{Sparse Implementation}

The published operator is implemented as a SciPy CSR sparse matrix multiplication in \texttt{tools/ppmi\_graph\_diffuse.py}. Each graph row is divided by its PPMI weight sum before multiplication. The corrected experiment driver uses a cleaned equivalent and includes an automated acceptance gate: for seed 42 and the published defaults, the generated float32 vectors match \texttt{fairytales\_ri\_seed42\_ppmi\_topk50.json} bit for bit.

\section{Evaluation}\label{sec:evaluation}

\subsection{Word Analogy Test}

Evaluation uses the Google Semantic-Syntactic Word Relationship test set~\cite{mikolov2013analogies}. For an analogy tuple $(a, b, c, d)$, the predicted fourth word is
\[
\hat{d} = \arg\max_{w \notin \{a,b,c\}} \cos(\mathbf{v}(w), \mathbf{v}(b) - \mathbf{v}(a) + \mathbf{v}(c)).
\]
A prediction is counted as correct when $\hat{d} = d$. Analogies containing any out-of-vocabulary word are discarded. For fairytales, the headline metric is the covered family-category semantic subset because it is the only covered semantic category. Fairytales syntactic coverage exists, but existing outputs place the selected seed42 RI+PPMI run at only 1.4\% on 2,006 valid syntactic questions, so syntactic accuracy is treated as a limitation check rather than as the positive result. We report both semantic and syntactic accuracy on text8 because syntactic coverage is large enough to provide a higher-power comparison.

\subsection{Word Similarity Benchmarks}

SimLex-999~\cite{hill2015} consists of 999 word pairs with human annotations of strict semantic similarity. WordSim-353~\cite{finkelstein2001} consists of 353 pairs annotated for broader semantic relatedness. For both benchmarks, performance is measured by the Spearman rank correlation between human scores and cosine similarities induced by the embeddings.

\subsection{Baselines}

All methods use 200-dimensional vectors. Table~\ref{tab:baselines} summarizes the settings.

\section{Results}

The positive result is confined to the Google benchmark's family section. Of its 506 questions, 272 are valid for every fairytales embedding and seed. The primary uncertainty estimates use a nested bootstrap that resamples the five paired seeds and then resamples aligned questions. Table~\ref{tab:family_decomposition} reports the endpoint estimates, and Table~\ref{tab:paired_effects} isolates the factors. The separate 40-question general grid is reported alongside the family analysis but is not pooled with it because the question sets differ.

\begin{table}[!t]
\footnotesize
\centering
\caption{Baseline methods and reported context windows. Graph-processed rows use a graph window of 8 unless otherwise noted. SVD denotes Singular Value Decomposition.}
\label{tab:baselines}
\setlength{\tabcolsep}{3pt}
\begin{tabular}{@{}ll@{}}
\toprule
Method & Setting \\
\midrule
RI uniform & 200d, nnz=8, init window 10 \\
RI+PPMI & uniform RI; PPMI window 8, min count 2, $K=50$ \\
PPMI+SVD & 200d truncated SVD, window 10 \\
Binary+SVD & 200d binary SVD, window 10 \\
CBOW & Word2Vec, window 4, 50 epochs, seed 42 \\
Skip-gram & Word2Vec, window 8, 50 epochs, seed 42 \\
Bloom & 200-bit, $h=5$, top-20 neighbors, graph window 8 \\
\bottomrule
\end{tabular}
\end{table}

\subsection{Comparison of Embedding Methods}

Uniform RI obtains 19.41\% family accuracy across five seeds. The complete one-step PPMI pipeline obtains 30.74\%, for a paired gain of 11.32 percentage points [6.93, 15.89]. The seed-specific post-filter range is 28.3\% to 34.6\%; it is descriptive and is not used for model selection. The same seed-42 output reports 1.4\% syntactic accuracy on 2,006 valid syntactic questions. Table~\ref{tab:fairytales_main} retains the previously reported single-run comparisons and similarity values, but they are auxiliary observations rather than evidence for the corrected causal decomposition.

\begin{table}[!t]
\footnotesize
\centering
\caption{Five-seed endpoint accuracy on the 272/506 covered family questions. Intervals are nested-bootstrap 95\% confidence intervals.}
\label{tab:family_decomposition}
\setlength{\tabcolsep}{3.5pt}
\begin{tabular}{lcc}
\toprule
Setting & Accuracy (\%) & 95\% CI \\
\midrule
RI initialization & 19.41 & [15.23, 23.74] \\
RI + robust scaling & 22.65 & [18.11, 27.37] \\
RI + row norm & 19.41 & [15.38, 23.74] \\
One-step average + row norm & 25.59 & [20.84, 30.44] \\
One-step average + robust & 30.74 & [25.33, 36.26] \\
\bottomrule
\end{tabular}
\end{table}

\begin{table}[!t]
\scriptsize
\centering
\caption{Previously reported fairytales family accuracy and similarity checks before and after graph processing. The RI+PPMI row uses uniform RI and reports five-seed analogy means. Other post-processed rows are retained single-run auxiliary comparisons and are not part of the corrected decomposition. Similarity cells show raw/post Spearman $\rho$. SimLex uses 889/999 pairs and WordSim uses 225/353 pairs.}
\label{tab:fairytales_main}
\setlength{\tabcolsep}{1.2pt}
\renewcommand{\arraystretch}{0.82}
\begin{tabular}{@{}lccccc@{}}
\toprule
Method & Raw & Post & $\Delta$ & SimLex & WordSim \\
\midrule
RI legacy TF-IDF graph & 18.8 & 19.1 & +0.3 & -0.056/0.093 & 0.139/0.264 \\
RI+PPMI & 19.4$\pm$0.7 & 30.7$\pm$2.9 & +11.3 & -0.056/0.020 & 0.139/0.264 \\
PPMI+SVD & 26.8 & 19.9 & -6.9 & 0.195/0.070 & 0.439/0.234 \\
Binary+SVD & 29.0 & 19.5 & -9.5 & 0.077/0.075 & 0.254/0.320 \\
CBOW & 25.7 & 12.9 & -12.8 & 0.241/0.095 & 0.405/0.231 \\
Skip-gram & 22.1 & 19.1 & -3.0 & 0.215/0.107 & 0.400/0.278 \\
\bottomrule
\end{tabular}
\end{table}

\subsection{Bloom Filters versus Random Indexing}

The Bloom result is a negative result for the tested corrected configuration ($m=200$, $h=5$, top-20 neighbors), not a broad claim about all Bloom-style sketches. The performance gap is consistent with several possible factors: loss of frequency information under OR accumulation, hash collisions at moderate bit density, and distortion of cosine geometry after bipolar recoding. Isolating each factor would require a separate ablation.

\begin{table}[!t]
\footnotesize
\centering
\caption{Bloom filter versus Random Indexing on fairytales family analogies (\%). Graph processing uses the one-step PPMI operator. The RI row reports the five-seed mean; Bloom is a single corrected configuration.}
\label{tab:bloom_vs_ri}
\setlength{\tabcolsep}{4pt}
\begin{tabular}{lcc}
\toprule
Setting & Init. only & + graph average \\
\midrule
Corrected Bloom (top-20, $h=5$) & 12.9 & 10.7 \\
Random Indexing (uniform) & 19.4$\pm$0.7 & 30.7$\pm$2.9 \\
\bottomrule
\end{tabular}
\end{table}

\subsection{Confound Analysis}

The family decomposition separates rescaling from graph averaging. Robust scaling alone adds 3.24 points [1.25, 5.38]. With robust scaling removed from both endpoints, one-step averaging adds 6.18 points [2.63, 9.92]. Thus averaging contributes beyond rescaling on this family set. Each contrast pairs identical seed IDs and aligned questions; all five seeds cover the same 272 of 506 questions.

The general grid reaches the opposite conclusion. On its separate 40/40 questions, averaging without robust scaling changes accuracy by -6.50 points [-14.50, -0.50]. The full pipeline changes it by -6.00 points [-13.50, -0.50]. PPMI versus the paper-defined TF-IDF graph package changes it by -3.50 points [-10.00, 0.00], and matching the RI and graph windows changes it by 0.50 points [0.00, 3.00]. These graph packages differ in weighting, filtering, positive truncation, and pruning, so their contrast is not attributable to weighting alone. The results support a family-specific effect, not a general RI improvement.

\begin{table}[!t]
\footnotesize
\centering
\caption{Paired marginal effects in percentage points. All intervals are nested-bootstrap 95\% confidence intervals across five seeds and aligned questions.}
\label{tab:paired_effects}
\setlength{\tabcolsep}{2.5pt}
\begin{tabular}{lcc}
\toprule
Contrast & Effect & 95\% CI \\
\midrule
Family: robust scaling alone & +3.24 & [1.25, 5.38] \\
Family: averaging, no robust scaling & +6.18 & [2.63, 9.92] \\
Family: full pipeline versus raw & +11.32 & [6.93, 15.89] \\
General: robust scaling alone & -1.50 & [-6.50, 2.00] \\
General: averaging, no robust scaling & -6.50 & [-14.50, -0.50] \\
General: full pipeline versus raw & -6.00 & [-13.50, -0.50] \\
General: PPMI graph minus TF-IDF graph & -3.50 & [-10.00, 0.00] \\
General: matched minus mismatched windows & +0.50 & [0.00, 3.00] \\
\bottomrule
\end{tabular}
\end{table}

The older auxiliary normalization results are retained in Table~\ref{tab:normalization_ablation} for continuity. They combine single-run settings and operators and do not identify a marginal effect. No corrected claim relies on them.

\begin{table}[!t]
\footnotesize
\centering
\caption{Previously reported auxiliary normalization and graph checks (\%). These single-run settings, except the PPMI family mean, are not the corrected five-seed decomposition.}
\label{tab:normalization_ablation}
\setlength{\tabcolsep}{4pt}
\begin{tabular}{lcc}
\toprule
Setting & Fairytales & Text8 sub2m \\
\midrule
Initialization only & 18.8 & 5.3 \\
+ Robust scaling only & 21.3 & 11.7 \\
+ Legacy graph pass + $\ell_2$ only & 14.3 & 2.0 \\
+ Legacy TF-IDF graph + robust & 19.1 & 5.3 \\
+ Legacy TF-IDF residual control & 26.1 & -- \\
+ PPMI top-$K=50$ ($\alpha=0.3$) & 30.7$\pm$2.9 & 12.6 \\
\bottomrule
\end{tabular}
\end{table}

\begin{table}[!t]
\scriptsize
\centering
\caption{Text8 sub2m analogy accuracy (\%) for selected baselines and the RI+PPMI pipeline. Valid columns report the number of evaluated Google analogy questions covered by each vocabulary.}
\label{tab:text8_baselines}
\setlength{\tabcolsep}{2.2pt}
\begin{tabular}{lcccc}
\toprule
Method & Sem. $n$ & Sem. & Syn. $n$ & Syn. \\
\midrule
RI raw & 506 & 5.3 & 8,946 & 0.7 \\
RI+PPMI & 506 & 12.6 & 8,946 & 1.4 \\
PPMI+SVD & 506 & 15.8 & 8,946 & 5.0 \\
Binary+SVD & 506 & 6.7 & 8,946 & 2.6 \\
CBOW & 380 & 41.1 & 7,700 & 9.0 \\
Skip-gram & 380 & 24.2 & 7,700 & 3.7 \\
\bottomrule
\end{tabular}
\end{table}

On text8, the reported RI+PPMI result is not competitive with the single-run neural baselines. CBOW and Skip-gram have higher semantic accuracy, though their 380 valid semantic questions differ from the 506 questions covered by RI and SVD. RI+PPMI reaches 1.4\% syntactic accuracy on 8,946 covered questions. These previously reported text8 values are retained, but they are not part of the new five-seed decomposition.

\subsection{Held-Out Alpha Selection and Similarity}

The corrected alpha sweep uses 17 held-out questions with no overlap with the reported family or general sets. Across five seeds, the tested means for $\alpha=\{0,0.15,0.3,0.5,0.7,1\}$ are 44.71\%, 48.24\%, 52.94\%, 23.53\%, 7.06\%, and 0.00\%. The highest tested mean is at $\alpha=0.3$, with a 95\% interval of [31.76, 74.12]. The $\alpha=0$ artifacts match the initialization-only artifacts byte for byte for all five seeds.

The retained similarity results in Table~\ref{tab:fairytales_main} show that strict similarity and broader relatedness do not move uniformly under graph processing. The seed-42 RI+PPMI artifact has SimLex $\rho=0.020$ and WordSim $\rho=0.264$. These existing similarity values are descriptive and were not used for hyperparameter selection in the corrected analysis.

\subsection{Runtime Comparison}

Table~\ref{tab:runtime} reports wall-clock training time for each method on the fairytales corpus. All methods were timed on the same machine (RTX 5060, 8 GB). The RI+PPMI time was measured using the actual pipeline behind the reported result: uniform Random Indexing initialization followed by PPMI graph construction and one residual averaging step with $K=50$, $\alpha=0.3$, and forced graph-cache recomputation.\footnote{A reproducibility detail is that SVD and RI initialization runs used window 10 following their defaults, while graph construction used window 8. This window difference affects co-occurrence density and should be noted when interpreting runtime and accuracy comparisons.}

\begin{table}[!t]
\footnotesize
\centering
\caption{Wall-clock training time on fairytales corpus. Accuracy reports raw/pre-processing family accuracy for consistency; the RI+PPMI final mean is 30.7$\pm$2.9\%.}
\label{tab:runtime}
\setlength{\tabcolsep}{4pt}
\begin{tabular}{lcc}
\toprule
Method & Time (s) & Sem. Acc. (\%) \\
\midrule
CBOW & 33 & 25.7 \\
Binary+SVD & 41 & 29.0 \\
PPMI+SVD & 42 & 26.8 \\
RI + PPMI pipeline & 52 & 19.4$\pm$0.7 \\
Skip-gram & 105 & 22.1 \\
\bottomrule
\end{tabular}
\end{table}

These timings do not support a speed claim. CBOW, Binary+SVD, and PPMI+SVD are faster than the RI+PPMI pipeline under these runs, while Skip-gram is slower. The comparison is also confounded by window choices, so it should not be used as a clean cross-method accuracy comparison. The controlled general grid does not establish an advantage for the PPMI graph package over the TF-IDF package.

\subsection{Memory Analysis}

The best fairytales PPMI graph contains 279,150 directed edges after top-$K=50$ pruning. With $|\mathcal{V}|=18{,}254$ and $d=200$, a single float32 embedding matrix requires about 14.6 MB, while a float64 working matrix requires about 29.2 MB. A CSR graph with float64 weights and 32-bit indices requires about 3.4 MB for edge values, column indices, and row pointers. The core working state is therefore roughly 62 MB for two float64 embedding matrices plus the graph, or about 91 MB if an additional accumulator matrix is materialized. The JSON files used in the experiments are larger because they store decimal text for reproducibility; they are not required by the algorithm itself. The result establishes the sparse working-state scale of RI+PPMI. We do not report a sparse versus sparse memory ratio against PPMI+SVD because the experiment logs do not contain measured peak memory for that baseline.

\section{Limitations}

The positive result is narrow. The family analysis covers 272 of 506 kinship questions, all from one corpus and one benchmark relation. It does not support claims about broad semantics, general RI quality, or downstream usefulness. Fairytales syntactic coverage is larger at 2,006 valid questions, but the selected seed-42 RI+PPMI artifact reaches only 1.4\% syntactic accuracy. The corrected $\alpha$ sweep uses a disjoint 17-question development set, but the PPMI graph definition and other design choices were inherited from the original exploratory study rather than selected in a fully independent preregistered protocol.

The reported RI+PPMI configuration mixes context windows: RI initialization uses window 10 and the PPMI graph uses window 8. In the general grid, matching the windows changes accuracy by only 0.50 points [0.00, 3.00], but that result comes from a different 40-question set and cannot settle the family-specific window effect. The same general grid finds negative effects for the complete pipeline and for averaging without robust scaling. On text8, the retained single-run RI+PPMI result is below CBOW and Skip-gram, and valid-question counts differ across vocabularies. The reported SimLex value is also weak. The Bloom result covers only one corrected configuration.

The source of the headline checkpoint is \texttt{tools/ppmi\_graph\_diffuse.py}. The canonical \texttt{iterative\_vectors\_v3.py} path uses a different update and must not be used to reproduce the 30.7\% number. The corrected driver tests a bit-for-bit match against the seed-42 published artifact, but the distinction between these code paths is an important provenance limitation. Older auxiliary tables contain single-run results from other graph-processing paths.

\section{Conclusion}

A sparse embedding pipeline based on uniform RI, global PPMI graph construction, one residual graph average, and terminal robust normalization was examined. It avoids dense co-occurrence matrix materialization, dense factorization, and gradient training, but it still constructs sparse global co-occurrence statistics and a sparse graph operator.

First, Bloom filter sketches underperform Random Indexing in the tested corrected configuration. The performance difference is consistent with loss of frequency information under OR accumulation and distortion of cosine geometry after bipolar recoding, but this mechanism remains a hypothesis because the factors were not isolated.

Second, on the 272 covered fairytales family questions, the complete pipeline improves accuracy by 11.32 points [6.93, 15.89] across five paired seeds. Averaging without robust scaling contributes 6.18 points [2.63, 9.92], so the family-set improvement is not attributable to rescaling alone.

Third, the separate general grid contradicts a broad improvement claim: the complete pipeline changes accuracy by -6.00 points [-13.50, -0.50], and averaging without robust scaling changes it by -6.50 points [-14.50, -0.50]. The supported conclusion is therefore limited to kinship analogies in this fairytales setting. The retained text8, similarity, and single-run baseline checks also provide no basis for a general competitiveness claim. Future work should test the family-specific finding on independent corpora, relations, and downstream tasks.

\section*{Code Availability}

Code, preprocessing scripts, and experiment drivers are available at \url{https://github.com/ShadowAsura/Word-Embeddings-and-Bloom-Filters}.


\begin{thebibliography}{99}

\bibitem{mikolov2013}
T.~Mikolov, K.~Chen, G.~Corrado, and J.~Dean.
\newblock Efficient estimation of word representations in vector space.
\newblock In \textit{ICLR Workshop}, 2013.

\bibitem{pennington2014}
J.~Pennington, R.~Socher, and C.~D.~Manning.
\newblock GloVe: Global vectors for word representation.
\newblock In \textit{EMNLP}, 2014.

\bibitem{baroni2014}
M.~Baroni, G.~Dinu, and G.~Kruszewski.
\newblock Don't count, predict! A systematic comparison of context-counting vs. context-predicting semantic vectors.
\newblock In \textit{ACL}, 2014.

\bibitem{levy2015}
O.~Levy, Y.~Goldberg, and I.~Dagan.
\newblock Improving distributional similarity with lessons learned from word embeddings.
\newblock \textit{Transactions of the Association for Computational Linguistics}, 3:211--225, 2015.

\bibitem{hill2015}
F.~Hill, R.~Reichart, and A.~Korhonen.
\newblock SimLex-999: Evaluating semantic models with (genuine) similarity estimation.
\newblock \textit{Computational Linguistics}, 41(4):665--695, 2015.

\bibitem{finkelstein2001}
L.~Finkelstein, E.~Gabrilovich, Y.~Matias, E.~Rivlin, Z.~Solan, G.~Wolfman, and E.~Ruppin.
\newblock Placing search in context: The concept revisited.
\newblock In \textit{WWW}, 2001.

\bibitem{mikolov2013analogies}
T.~Mikolov, W.-t.~Yih, and G.~Zweig.
\newblock Linguistic regularities in continuous space word representations.
\newblock In \textit{NAACL-HLT}, 2013.

\bibitem{pedregosa2011}
F.~Pedregosa, G.~Varoquaux, A.~Gramfort, V.~Michel, B.~Thirion, O.~Grisel, M.~Blondel, P.~Prettenhofer, R.~Weiss, V.~Dubourg, J.~Vanderplas, A.~Passos, D.~Cournapeau, M.~Brucher, M.~Perrot, and E.~Duchesnay.
\newblock Scikit-learn: Machine learning in Python.
\newblock \textit{Journal of Machine Learning Research}, 12:2825--2830, 2011.

\bibitem{kanerva2000}
P.~Kanerva, J.~Kristoferson, and A.~Holst.
\newblock Random indexing of text samples for latent semantic analysis.
\newblock In \textit{Cognitive Science Society}, 2000.

\bibitem{faruqui2015}
M.~Faruqui, J.~Dodge, S.~K.~Jauhar, C.~Dyer, E.~Hovy, and N.~A.~Smith.
\newblock Retrofitting word vectors to semantic lexicons.
\newblock In \textit{NAACL-HLT}, 2015.

\bibitem{mu2018}
J.~Mu and P.~Viswanath.
\newblock All-but-the-top: Simple and effective postprocessing for word representations.
\newblock In \textit{ICLR}, 2018.

\bibitem{levy2014}
O.~Levy and Y.~Goldberg.
\newblock Neural word embedding as implicit matrix factorization.
\newblock In \textit{NIPS}, 2014.

\bibitem{serra2018}
E.~Serra and A.~Karatzoglou.
\newblock Getting deep recommenders fit: Bloom embeddings for sparse binary input/output.
\newblock In \textit{RecSys}, 2018.

\bibitem{svenstrup2017}
D.~Svenstrup, J.~M.~Hansen, and O.~Winther.
\newblock Hash embeddings for efficient word representations.
\newblock In \textit{NIPS Workshop}, 2017.

\bibitem{sahlgren2005}
M.~Sahlgren.
\newblock An introduction to random indexing.
\newblock In \textit{Methods of Information Extraction and Retrieval}, 2005.

\bibitem{bloom1970}
B.~H.~Bloom.
\newblock Space/time trade-offs in hash coding with allowable errors.
\newblock \textit{Communications of the ACM}, 13(7):422--426, 1970.

\bibitem{zhu2024}
R.-J.~Zhu, Y.~Zhang, E.~Sifferman, T.~Sheaves, Y.~Wang, D.~Richmond, P.~Zhou, and J.~K.~Eshraghian.
\newblock Scalable matmul-free language modeling.
\newblock \textit{arXiv preprint arXiv:2406.02528}, 2024.

\bibitem{halko2011}
N.~Halko, P.~G.~Martinsson, and J.~A.~Tropp.
\newblock Finding structure with randomness: Probabilistic algorithms for constructing approximate matrix decompositions.
\newblock \textit{SIAM Review}, 53(2):217--288, 2011.

\end{thebibliography}
\end{document}